\documentclass[11pt]{article}
\usepackage{scl}
\usepackage{times}
\usepackage{url}
\usepackage{latexsym}
\usepackage{array}
\usepackage{subfig}
\usepackage[final]{graphicx}
\usepackage{multirow}
\usepackage{wrapfig}
\usepackage{graphicx}
\usepackage{float}
\usepackage{dirtytalk}
\usepackage{booktabs}
\usepackage[prependcaption,textsize=tiny]{todonotes}

\newcommand*{\affaddr}[1]{#1} 

\newcommand*{\affmark}[1][*]{\textsuperscript{#1}}
\newcommand*{\email}[1]{\texttt{#1}}

\usepackage[utf8x]{inputenc}

\title{Evaluating Neural Word Embeddings for Sanskrit}

\author{Jivnesh Sandhan\affmark[1],
\textbf{Om Adideva\affmark[2],}
\textbf{Digumarthi Komal\affmark[1],}\\
\textbf{Laxmidhar Behera,\affmark[1,3]}
\textbf{and Pawan Goyal\affmark[4]}  
\\

\affaddr{\affmark[1]Dept. of Electrical Engineering, IIT Kanpur,}\\
\affaddr{\affmark[2]Vellore Institute of Technology,}
\affaddr{\affmark[3]Tata Consultancy Services,}\\
\affaddr{\affmark[4]Dept. of Computer Science and Engineering, IIT Kharagpur}\\
\email{jivnesh@iitk.ac.in, pawang@cse.iitkgp.ac.in}\\}

\date{}

\begin{document}
\maketitle
\begin{abstract}
Recently, the supervised learning paradigm's surprisingly remarkable performance has garnered considerable attention from Sanskrit Computational Linguists. As a result, the Sanskrit community has put laudable efforts to build task-specific labeled data for various downstream Natural Language Processing (NLP) tasks. The primary component of these approaches comes from representations of word embeddings.
Word embedding helps to transfer knowledge learned from readily available unlabelled data for improving task-specific performance in low-resource setting. Last decade, there has been much excitement in the field of digitization of Sanskrit. To effectively use such readily available resources, it is very much essential to perform a systematic study on word embedding approaches for the Sanskrit language.
In this work, we investigate the effectiveness of word embeddings. We classify word embeddings in broad categories to facilitate systematic experimentation and evaluate them on four intrinsic tasks.  We investigate the efficacy of embeddings approaches (originally proposed for languages other than Sanskrit) for Sanskrit along with various challenges posed by language.
\end{abstract}

\section{Introduction}
\label{intro}
The supervised learning paradigm has shown remarkable performance in many downstream Natural Language Processing (NLP) tasks. In this domain, the NLP community has witnessed the emergence of neural-based approaches with the state of the art performance.  The dramatic popularity of such approaches has garnered considerable attention from Sanskrit Computational Linguistics (SCL) community. The primary prerequisite for the feasibility of such approaches is the availability of task-specific labeled data. Therefore, as a first step, there was a need for preparing such task-specific resources. As a result, laudable efforts have been put by the community for building task-specific labeled data for various NLP tasks \cite{krishna-etal-2017-dataset,hellwig-nehrdich-2018-sanskrit,hellwig-etal-2020-treebank}. The availability of such resources triggered the emergence of the data-driven neural-based approaches into the SCL field. The wide success of neural-based data-driven approaches is greatly justified by the state of the art performance for many downstream tasks for Sanskrit, namely, segmentation \cite{hellwig-nehrdich-2018-sanskrit,reddy-etal-2018-building,krishna-etal-2018-free}, dependency parsing \cite{amrith21,krishna2020neural,sandhan2021little,krishna-etal-2020-keep}, semantic type identification \cite{sandhan-etal-2019-revisiting,krishna-etal-2016-compound}, word order linearisation \cite{krishna-etal-2019-poetry,amrith21} and morphological parsing \cite{gupta-etal-2020-evaluating,krishna-etal-2018-free}. 

The supervised approaches have shown remarkable performance in several downstream NLP tasks and many of these use word embeddings as a primary component.
In 2013, \newcite{mikolov2013distributed} proposed a new milestone in the field of word embedding, which completely revolutionized Natural Language Processing (NLP). From there, word embedding has been a buzzword in NLP and has become part and parcel of downstream applications.  These word embedding approaches adopt a semi-supervised learning paradigm. They do not depend on any explicitly labeled data or human supervision.  Availability of massive unlabelled corpus and ease of integration into downstream NLP tasks made them exciting for the researchers. Sanskrit, the cultural heritage of India, has 30 million extant manuscripts \cite{goyal-etal-2012-distributed}. Last decade, there has been much excitement in digitization for Sanskrit, as a result, we have DCS\footnote{\url{http://www.sanskrit-linguistics.org/dcs/index.php}}, The Sanskrit Library\footnote{\url{https://sanskritlibrary.org/}} and GRETIL\footnote{\url{http://gretil.sub.uni-goettingen.de/gretil.html}}. To reduce the labeled data dependency, it's crucial to take advantage of such cheaply available resources. 

This motivates us to systematically investigate word embedding approaches that are originally proposed for resource-rich languages like English. Most of the research in NLP disproportionately focuses on resource-rich languages due to readily available language resources: digitized texts and gold standard labeled data \cite{joshi-etal-2020-state}. Also, the recent word embedding approaches have been studied for resource-rich languages such as English. However, such analysis is lacking for Sanskrit. 
In addition to the morphological rich nature of language, the Sanskrit is a relatively free word order language, and most words are compound words that demand special attention for modeling the word embedding approach. To conduct a systematic study on the efficacy of these approaches, we categorize word embeddings into two categories, namely, \textit{static} and \textit{contextualized}. The \textit{static} embeddings generate a single representation for all the sense of meanings possible for a word, irrespective of surrounding context words. In contrast to that \textit{contextualized} embeddings generate a separate representation for each sense of the meaning depending on the surrounding context words. This work provides fertile soil for investigating the following questions: (1) Which linguistic phenomenons are captured by word embeddings? (2) Out of existing word embedding approaches, which embeddings are more suitable for Sanskrit? (3) What are the shortcomings of existing approaches specific to Sanskrit?
Based on our experiments, our major findings are as follows:
\vspace{-1mm}
\begin{enumerate}\itemsep0em
    \item Our intrinsic evaluation results illustrate that word embedding models trained on Sanskrit corpus are able to capture exceptionally good amount of the syntactic information. However, they perform relatively poorly with the semantic notion of relatedness - similarity mainly due to Out of vocabulary problem.
    \item The \textit{contextualized} word embedding model by \newcite{elmo} outperforms all the word embeddings in all the intrinsic tasks with a large margin except semantic categorization task (Section \ref{experiments}). We employ contextual models without a context to measure efficacy of these approaches compared to static approaches. Applying \textit{contextualized} models without context is similar to just using their underlying static word representation.
 
    \item The qualitative analysis shows that the \textit{contextualized} representations generated by \newcite{elmo} and \newcite{ALBERT:} for polysemous words indicating the same sense of meaning forms a cluster for each sense of meaning (Section \ref{context_cluster}). This is indeed a serendipitous outcome for such data-hungry models when trained on negligibly small amount of corpus available for Sanskrit.
    
    \item We release our codebase, datasets and models publicly at: \url{https://github.com/jivnesh/EvalSan}. This might help the research community to do further analysis and apply these embeddings to downstream NLP applications of Sanskrit.
\end{enumerate}

%
%
    %
    %

\section{Neural Word Embedding}
Neural word embedding methods construct vector space so that semantically related words are projected in closer vicinity compared to unrelated words. Word embedding is a dense representation of words in low-dimensional vector space. It mainly exploits the distribution hypothesis: neighboring words share similar meaning \cite{firth1957synopsis,harris1954distributional}. 

\begin{table}[!h]
\centering
\resizebox{0.6\textwidth}{!}{%
\begin{tabular}{|l|l|l|}
\hline
\textbf{Category}  & \textbf{Input level} & \textbf{Models}    \\ \hline
Static   & Token & Word2vec, GloVe    \\ \hline
   & Subword & FastText    \\ \hline
   & Character & CharLM    \\ \hline \hline

Contextualized  & Subword & ALBERT   \\ \hline 
 & Character & ELMo    \\ \hline
\end{tabular}%
}
\caption{Categorization of embedding models}
\label{tab:introTable}
\end{table}

As shown in Table \ref{tab:introTable}, we consider the two broader categories of embedding approaches, namely, \textit{static} and \textit{contextualized}. Further, we divide them based on an atomic unit of input fed to the model, i.e., token level, subword level and character level. First, we consider token level embeddings: \texttt{word2vec} \cite{mikolov2013distributed} and \texttt{GloVe} \cite{pennington-etal-2014-glove}. We consider \texttt{word2vec} as a strong baseline because it is the basic backbone of all the neural-based follow-up works in this domain. While \texttt{word2vec} focuses on the local context of a target word, \texttt{GloVe} is an alternative approach that considers the global context of a target word. Therefore, \texttt{GloVe} can be seen as representative of count-based models; hence we choose to include it in our study.  However, these token-based approaches suffer from two serious issues: (1) Out of vocabulary (OOV) problem: they fail to obtain representation for words which are not seen during training (2) Polysemy problem: a single representation for multiple meanings of words. In order to handle the first drawback, sub-word \cite{wieting-etal-2016-charagram,bojanowski-etal-2017-enriching,heinzerling-strube-2018-bpemb} and character level \cite{kim_charLM,jozefowicz2016exploring} compositional modeling is proposed. These models learn composition functions to obtain word-level representation of a word using character level or subword level modeling. In the next category, the advent of contextualized embeddings is attributed to the polysemy problem.
The year 2018 is an inflection point in the field of NLP with the launching of contextualized embedding by \newcite[\texttt{ELMo}]{elmo}. The specific meaning of the word can be understood by the context words surrounded by it. Hence, in contextualized embedding approach, the representation of each token word is composed as a function of entire sentence as a input.
 The emergence of another contextual embedding by  \newcite[\texttt{BERT}]{bert} can be seen as a clever recipe of a bunch of excellent ideas bubbled in the field of NLP in recent years. 
 We consider a derivative of \texttt{BERT} known as \texttt{ALBERT} \cite{ALBERT:} which is light version of \texttt{BERT} with relatively less number of parameters. This momentous development enabled researchers to use this powerhouse as an off-the-shelf feature extractor: saving the training model's computational efforts from scratch.


\section{System Description}
\paragraph{Tokenizer:}
Token-based word embeddings such as \texttt{word2vec} and \texttt{GloVe} suffer from OOV problem, i.e., they are not capable of producing vector representation for words that are not seen during training. This problem is even more prominent for Sanskrit due to the following reasons: (1) morphological-rich nature (2) highly inflectional (3) most of the text is available in \textit{sandhied} format. Hence, token-based approaches are highly ineffective for Sanskrit. We take inspiration from \newcite{heinzerling-strube-2018-bpemb} and leverage \texttt{sentencepiece} tokenizer \cite{sentencepiece} to overcome OOV problem. The \texttt{sentencepiece} is unsupervised tokenizer used for neural-based tasks such as machine translation and sequence generation task. This is a purely data-driven tokenizer and does not require any pretokenization. Also, this is language independent approach where words are considered as a sequence of unicode characters. The \texttt{sentencepiece} overcomes the challenge as mentioned earlier by creating a new vocabulary instead of relying on real word boundaries. This method uses a greedy approach to figure out subword units that maximize the likelihood of the language model \cite{sentencepiece}. There exist other unsupervised tokenizers such as \texttt{Wordpiece} \cite{wordpiece} and \texttt{BPE} \cite{bpe}. However, these approaches expect tokenized words as input and most of the available text for Sanskrit is in \textit{sandhied} format. Hence, \texttt{sentencepiece} is the most befitting choice as a tokenizer. We demonstrate \texttt{sentencepiece} tokenization with an example as shown below.\footnote{Originally, we apply tokenization on SLP1 transliterated data. Just to illustrate, we show an example in the romanized (IAST) transliteration scheme.} Here, \texttt{sentencepiece} considers space character as a normal character and is replaced by \say{\_} character. Note that \textit{ manā\_bhava} is a word in our newly generated vocabulary, originally part of two words.
\begin{center}
   \textbf{Sandhied sentence (original)} \\
   \textit{man-manā bhava mad-bhakto mad-yājī māṁ namaskuru \\
mām evaiṣyasi yuktvaivam ātmānaṁ mat-parāyaṇaḥ} \\
\end{center}
\begin{center}
   \textbf{Segmented by \texttt{sentencepiece} tokenizer }\\
   \textit{\_man - manā\_bhava \hspace{0.1mm} \_mad -bhakto \hspace{0.1mm} \_mad - yājī \hspace{0.1mm} \_māṁ \hspace{0.1mm} \_namas kuru \\
\_mām \hspace{0.1mm} \_evai ṣyasi \hspace{0.1mm} \_yukt vaivam \hspace{0.1mm} \_ātmān aṁ \hspace{0.1mm} \_mat - parāyaṇ aḥ} 
\end{center}

\subsection{Static Embeddings}
\paragraph{Word2Vec:}
The work of \newcite{mikolov2013distributed} is marked as a new milestone in the field of word embeddings and popularly known as \texttt{word2vec}. This work was the basic backbone of the neural-based word embedding approach, which led to thousands of word embedding variants tackling various challenges. This model exploits the distributional semantics hypothesis: \say{semantically similar words are used in similar contexts} \cite{firth1957synopsis,harris1954distributional}. \texttt{Word2vec} often assumed as the first-class citizen of the deep learning models; however, technically, neither the architecture is deep, nor the model uses non-linearities. It adapts language modeling objectives to consider the additional context. Language modeling is the task where the next word is predicted given previous \textit{n} words. However, \texttt{word2vec} also considers next \textit{n} words as a context.  The basic intuition behind this model is to project the words into \textit{n} dimensional vector space such that semantically similar words are placed in closer vicinity than non-similar words. In other words, the dot product of similar (cosine similarity) words will be higher compare to that with non-similar words. In order to obtain geometric organization of words, authors propose two training methods: (1) Continuous Bag of Words \texttt{(CBOW)} (2) \texttt{Skip-gram}. While the \texttt{CBOW} uses context words to predict the target word, skip-gram predicts context words given the target word as an input. Note that context is considered as a window of \textit{n} words surrounding target word from forward and backward direction. Unlike prior sparse representations, these learned vector representations are dense vectors of dimension ranging from 50-1000 and these dimensions do not have clear semantic interpretation.

\paragraph{GloVe:} Prior to \newcite{mikolov2013distributed}, count-based models were prominent in the word embedding field. These count-based models  \cite{bengio2003neural,count_1,count_2} are explicitly designed to model the global statistics of a entire corpus. On the other hand, predictive models such as \texttt{word2vec} solely focus on the target word's local context. To gain the best of two worlds, \newcite{pennington-etal-2014-glove} proposed an alternative word embedding technique that integrates global statistics via matrix factorization and local context via the sliding window method. This approach focuses on the co-occurrence of each word with all other words in the entire corpus, therefore called Global Vectors \texttt{(GloVe)}. For the unsupervised models, the primary source of information to learn word representation is corpus statistics. However, how to exploit corpus statistics to generate word representations is an interesting problem. \newcite{pennington-etal-2014-glove} show that the ratio of co-occurrence probabilities of words A and B with C i.e. $P(A,C)/P(B,C)$  is more indicative of their relation than a direct co-occurrence probability i.e. $P(A,B)$ . Here, $P(A,B)$ is given by $X_{AB}/X_{A}$ where $X_{AB}$ is the number of times word $A$ co-occurs with word $B$ and $X_{A}$ is the total number of times word $A$ appears in the corpus. How can we use scalar quantity $P(A,C)/P(B,C)$ to learn meaningful vector representation of words?  \newcite{pennington-etal-2014-glove} propose a novel objective function to tackle this challenge.\footnote{We recommend the readers to refer original paper for derivation details of objective function.} The training objective of this model is to learn word vectors for words A and C such that their dot product gives a fair estimate of their co-occurrence counts. Formally, the objective can be expressed as follows: 
\[A^TC+b_{A}+b_{C}=log(X_{AC})\]
In the above equation, $b_{A}$ and $b_{C}$ are bias terms for the words A and C, respectively.  The generated vectors are considered as the final word embedding vectors for words A and C. Although, \texttt{GloVe} does not use neural network, \newcite{levy-etal-2015-improving} consider it as a predictive model due to following reason:  Unlike other count-based models, \texttt{GloVe} use the Stochastic Gradient Descent (SGD) optimization for non-convex objective function. The \texttt{GloVe} outperforms it's peer models (including \texttt{word2vec}) in several intrinsic and extrinsic evaluation tasks. The model works well with a small-sized corpus but is highly scalable to a larger corpus. Although the model demands much memory to populate the co-occurrence matrix in large corpora, the authors term it as a \say{one-time, up-front cost}.

\paragraph{FastText:}
Token-level embeddings like \texttt{word2vec} and \texttt{GloVe} suffer from \textit{Out of Vocabulary} (OOV) problem. That means these models cannot generate vector representation for words that are not present in vocabulary. For example, even if model learns embeddings for \textit{pācikā} and \textit{bhārya\d{h}}, it will not be able to handle the word \textit{pācikābhārya\d{h}}, if it is not a part of vocabulary. Hence, such approaches could not serve the need of morphologically rich languages. Also, the model is not capable of handling different possible inflections of a word (for example, \textit{rāmaḥ, rāmābhyām, rāmasya}) if such inflectional variants are not a part of the vocabulary. Hence, to overcome these shortcomings, \newcite[\texttt{FastText}]{bojanowski-etal-2017-enriching} proposed a new word embedding technique, specifically, for learning reliable representations for OOV words. To accomplish this, the authors propose to incorporate subword level information into the model. While learning vector representation of target word, simultaneously, representations of all n-grams (\textit{n} varies from 3 to 6) are also learned. For example, for n=4, the possible 4-grams of a word \textit{ajñānam} are \textit{<ajñ, jñāna, ñānam, ānam>}. The brackets are used for padding purposes to indicate the beginning and end of the word. With this modification, the regular SGNS algorithm is applied. This subword level representation learning empowers the model to generate a representation for OOV words by simply summing representations of n-grams corresponding to the target word.  \texttt{FastText} has been shown to improve performance on syntactic word analogy tasks, especially in the case of morphologically rich languages.

\paragraph{CharLM:}
\newcite[\texttt{CharLM}]{kim_charLM} proposed a new variant of language model that learns subword information through a character-level convolutional neural network (CNN). Language modeling is a task where the model predicts the next word given the \textit{n} previous context words. In a standard neural language modeling setting, the model takes word-level vector representation as input and predicts word-level information. This modeling decision makes the model blind to subword information. For example, the inflections of a root word should have structurally similar word embeddings because they shared common sub-strings except for affixes.  Therefore, to capture subword level information, \texttt{CharLM} takes a character-level input and learns composition function to generate word-level representation. This is achieved by extracting the features from CNN followed by a highway network. Finally, these extracted features are fed to a long short-term memory (LSTM) recurrent neural network language model to make word-level predictions. \newcite{kim_charLM} argues that without using word embeddings as input to the language model, their model gets the additional advantage of fewer parameters with competitive performance. Also, they show that the model is capable of encoding both semantic and orthographic information. They demonstrate that languages with rich morphology outperform word-level/morpheme-level LSTM baseline methods.

\subsection{Contextualized Embedding}
 Each word has multiple meanings and their uses highly depend on the context they are used. However, irrespective of context, these traditional approaches give a single vector representation for a word. Therefore, traditional word embeddings are called \textit{static} embeddings. For example, there are two possible meanings of a word \textit{cakrī}: (1) a potter (2) the holder of the disc and its uses highly depends on the context. In order to solve this polysemy problem, contextualized embeddings like \texttt{ELMo} \cite{elmo} and \texttt{BERT} \cite{bert} proposed. In literature, there exists two domains for applying language representations to downstream tasks, namely, \textit{feature-based} and \textit{fine-tuned}. The feature-based approach simply augments the learned representation into the task-specific model architecture and well-known example in this category \texttt{ELMo}. A fine-tuned approach such as \texttt{BERT}, integrates pretrained model architecture into task-specific model and trains all the parameters, including the pretrained component. The emergence of these embeddings is described as a beginning of a new era in a word embedding. Their exceptional popularity in the field launched thousands of contextualized embedding variants. Hence, we include \texttt{ELMo} and \texttt{ALBERT} (A Light \texttt{BERT}: derivative of \texttt{BERT} model with significantly less number of parameters) \cite{ALBERT:} in our study. From an architectural point of view, \texttt{ELMo} belongs to LSTM based language model family and \texttt{ALBERT} belongs to the transformer-based language model family. 

\paragraph{ELMo:}
The most compelling signs of \texttt{ELMo} in NLP are mainly attributed to ease of integration and stunning improvement in the state of the art performance for many Natural Language Understanding tasks. Unlike traditional \textit{static} word embeddings, \texttt{ELMo} computes a representation of each token word as a function of all the words in a sentence. Authors employ the language modeling objective; hence, these representations are also called as \texttt{ELMo} (Embeddings from Language Models). There are three salient features of \texttt{ELMo}: (1) \textit{contextual:} the word representation depends on surrounding context words (2) \textit{deep:} the word representation is weighted average of all the layers of model architecture  (3) \textit{character based:} character based modeling enables model to encode morphological clues to obtain robust representations for OOV words, especially, for languages with rich morphology.
Prior state of the art neural language models \cite{kim_charLM,jozefowicz2016exploring} takes context-independent token-level representation as an input to forward language modeling objective where next word is predicted based on previous \textit{n} words. Similarly, the backward language model predicts a word given \textit{n} following words, i.e., precisely opposite to the forward model. \texttt{ELMo} coupled both forward and backward language model objective function together.
 Through ablations and intrinsic evaluation experiments, authors discover that higher-level layers in model architecture capture semantic properties and lower-level layers are syntactic. They show that a linear combination of internal representation helps obtain a rich representation, which is a better decision over just using top layer representation. Simultaneously providing enriched representation via a weighted linear combination of all layers, the model can choose the most relevant features for specific downstream tasks.

\paragraph{ALBERT:}
 \texttt{BERT} is blend of several clever ideas that has been emerging in the field of NLP such as semi-supervised learning \cite{dai2015semisupervised}, \texttt{ELMO} \cite{elmo}, \texttt{ULMFIT} \cite{ulmfit} and \texttt{Transformer} \cite{attention_vaswani}.
The foundational concept used in \texttt{BERT} is a transformer architecture popularized by \newcite{attention_vaswani}. Inspired from \newcite[\texttt{ULMFIT}]{ulmfit}, \texttt{BERT} adopts the fine-tuning approach to augment learned contextual representation in downstream applications. Authors argue that the previous language representations are unidirectional language modeling objective and this choice limits the model to learn effective representations. First, \texttt{ELMo} from the LSTM family demonstrated the effectiveness of a bidirectional language modeling for learning enriched representations. Suppose we introduce bidirectional language modeling into a transformer with a vanilla language modeling objective. In that case, the problem will become trivial and it will not be possible to learn robust representations. Therefore, to tackle this challenge, authors adapt a vanilla language modeling objective to a masked language objective. Here, instead of predicting the next word from previous \textit{n} words, the model masks a few words from a sentence and learns to recover masked words from the remaining words. This is also called a denoising objective. In this way, \texttt{BERT} can take advantage of the surrounding context from both sides simultaneously. 

This monstrous model reports the state of the art results in many downstream tasks on the leaderboard at the expense of huge parameters. Despite the great success of \texttt{BERT}, it is highly impractical to train such a monstrous model with limited available resources for Sanskrit. For example, \texttt{BERT-base} has 110 million trainable parameters, making it computationally very intensive and nearly impossible to train on consumer devices. Therefore, we choose a derivative of \texttt{BERT} called as \texttt{ALBERT}, a light \texttt{BERT}, which has significantly few parameters than traditional \texttt{BERT} model with significantly better performance. The authors achieved this by making three key changes to the original \texttt{BERT}'s architecture. First, they factorize the large vocabulary embedding matrix used in \texttt{BERT} into two smaller ones such that dependency on the size of hidden layers is obliviated from the size of vocabulary embeddings. Second, they apply cross-layer parameter sharing. This way, the parameters are shared for similar sub-segments, thus ensuring that parameters are learned only once and are then reused for the subsequent blocks. This significantly reduces the number of trainable parameters in the model. Lastly, they propose a new inter-sentence coherence task called Sentence Order Prediction, instead of using original \texttt{BERT}'s Next Sentence Prediction (NSP). NSP is a binary classification loss that was specifically created to improve performance on downstream NLP tasks. In summary,  the \texttt{ALBERT} model is found to have 18x fewer parameters than \texttt{BERT-large} and trains 1.7x faster while achieving the state of the art results on many downstream tasks. This model is an important breakthrough as it brings the capabilities of \texttt{BERT} to a more practical and usable form that can be useful for real-world applications.      

\section{Intrinsic Evaluation}
To evaluate the quality of the embedding model, two primary methods are proposed in literature: (1) Intrinsic evaluation (2) Extrinsic evaluation.  Intrinsic evaluation refers to the class of methods to measure the quality of vector space without evaluating them on downstream tasks. For example, semantic similarity is the most traditional method to evaluate meaning representations intrinsically. 
On the other hand, extrinsic evaluation refers to the class of methods to assess the quality of vector representation when fed as an input to the machine learning-based model in the downstream NLP task. The wide range of NLP tasks that deals with lexical semantics such as chunking, part of speech tagging,  semantic role labeling and named entity recognition can be used for the extrinsic evaluation. In this work, we restrict our study to intrinsic tasks.  
Intrinsic evaluation directly checks for syntactic and semantic properties of words \cite{mikolov2013distributed,baroni-etal-2014-dont}. These tasks rely on query inventory datasets consisting of a query word and semantically related target word. These query inventories are readily available for resource-rich languages but not for Sanskrit. We create such inventories for four different intrinsic tasks: relatedness, synonym detection, analogy prediction, and concept categorization. For creating such query inventory dataset for these intrinsic evaluation tasks, we use \textit{Amarako\d{s}a}\footnote{\url{https://sanskrit.uohyd.ac.in/scl/amarakosha/index.html}} \cite{inproceedings_nair}, Sanskrit WordNet\footnote{\url{https://www.cfilt.iitb.ac.in/wordnet/webswn/index.php}} \cite{Kulkarni2017} and Sanskrit Heritage Reader\footnote{\url{https://sanskrit.inria.fr/DICO/reader.fr.html}} \cite{goyaldesign16,heut_13}. In extrinsic evaluation, feature extracted by word embeddings is fed as an input to downstream NLP task and corresponding performance metric is used as an indicator to compare across different embedding approaches. There is an implicit assumption that universal ranking of word embedding quality and high ranking embedding will enhance the downstream task performance. Although this family of evaluation methods gives signal on the relative strength of different word embedding approaches, it needs not be considered a general proxy for overall quality.

   \paragraph{Relatedness: }
  It is essential to distinguish between the notion of relatedness and synonyms \cite{agirre-etal-2009-study}. Synonym words share many semantic properties and they can be substituted with each other without changing the meaning of the sentences. On the other hand, semantically related words can co-occur in the same context or document and share any one semantic relationship such as meronymy or antonymy. Unlike synonym words, we can not substitute related words in place of each other. The notion of semantic relatedness between word pairs is one of the fascinating concepts in lexical semantics with a variety of NLP applications \cite{pilehvar2020embeddings}. This task is considered as an in-vitro evaluation framework for judging the quality of word embeddings.
  The word pairs are annotated based on the degree of relatedness on a numerical scale by human annotators for evaluating semantic relatedness. For example, the distance between the \textit{paricārikā} and \textit{vaidya} in the numerical scale from 0 to 3 could be 2.5 because both words belong often found together. Similarly, the human judgment score for word pair \textit{prādhyāpaka - karkaṭī} could be 0.5 because these two words do not share any semantic property.
  The word embedding performance is then evaluated by assessing the correlation between the average scores assigned by annotators and the cosine similarities between corresponding vector representations. 
  
  For the construction of such human-annotated test data for the new target language, a standard methodology is followed in the literature \cite{camacho-collados-etal-2015-framework}. It is divided into two steps: (1) English word pairs of a standard English dataset are translated into the target language (2)  language experts annotate these word pairs. For the English language, the dataset popularized by \newcite[WordSim353]{hassan_semantic} has 353 word pairs with a human relatedness score. However, this dataset is criticized due to the human judgment criterion, which conflates similarity and relatedness. Another alternative is RG-65 \cite{RG_65} dataset. However, we find that many words such as \textit{automobile, cushion, autograph etc.} can not be naturally translated into Sanskrit. Therefore, we use \textit{Amarako\d{s}a} to get word pairs due to its systematic organization of semantically related words. Here,  synonymous words are categorized into synsets and semantically related words into \textit{varga}. Therefore, we decide to select the word pairs from \textit{Amarako\d{s}a}. This choice obliviates the need for translation and annotation of word pairs into Sanskrit. We exploit systematic categorization based on relatedness presents in the \textit{Amarako\d{s}a}. We transform the relatedness task into a classification task where the task is to categorize word pairs into three classes based on cosine similarity scores between vector representations generated by corresponding embeddings. For selecting word pairs, we follow the following procedure. First, we sample 13,500 word pairs in such a way that an equal proportion of word pairs are selected from the following classes : (1) word pairs from the same synset (2) word pairs from same \textit{varga} but not from the same synset (3) and the remaining word pairs such that each word from a word pair belongs to different \textit{varga}.
  Out of 13,500, we use 4,500 word pairs as the development and remaining word pairs in the test set.  We use a development set to find out cosine similarity thresholds to classify word pairs into respective classes. Here, we use the macro level F-score as an evaluation metric to measure the performance.
  	 		 		 		 	
 \paragraph{Synonym Detection:} 
 Synonym detection \cite{baroni-etal-2014-dont,bakarov2018survey} is a task where the goal is to find the synonym of a given target word from the four synonym candidates. In this multiple-choice setting, a single answer is correct. For example, for target \textit{sūryaḥ}, synonym candidates are \textit{ādityaḥ} (correct), \textit{cañcalā}, \textit{candraḥ} and \textit{ākāśa}.  Here, we choose an option as a correct answer which shows the highest cosine similarity with the target word. We use accuracy as a metric to measure performance. We leverage \textit{Amarako\d{s}a} \cite{inproceedings_nair} synsets to build MCQs. It has 4,056 synsets categorized into 25 \textit{vargas}. Out of 4,056, only 2,581 synsets have more than one element. We choose a word pair from each such synset, assign one word as the target word and another word as the correct answer. Then, for the remaining options, we use three words from the same \textit{varga} but not from the synset where the target word belongs. The primary motivation of such a choice is that words belonging to the same \textit{varga} are semantically related and picking such related words as synonym candidates make the task more interesting and challenging. Our test data has  3,034 MCQs. We consider accuracy as an evaluation metric for this task.
 
 \paragraph{Analogy Prediction: } 
  The analogy prediction task popularized by \newcite{mikolov2013distributed}, a question is framed with word pairs exhibiting similar sense. For example, \say{If \textit{kapi} is similar to \textit{kapibhyām} then in the same sense \textit{yati} is similar to \textit{...}}
 Authors propose that a simple algebraic operations suffice to answer these analogy questions. In above example, in order to find similar word for \textit{yati}, we need to calculate \textit{X = vector(\say{kapibhyām}) − vector(\say{kapi}) +
vector(\say{yati})}. Then, we find the nearest neighbour to X as per cosine similarity metric and use it as answer to question. Note that we do not use input words while calculating nearest neighbour.
 \begin{table}[h]
 
\centering
\begin{tabular}{ccccc}
\cmidrule(r){1-5}
 Relationship&\multicolumn{2}{c}{Word Pair 1}
&
\multicolumn{2}{c}{Word Pair 2} \\
\hline
declension&kapi& kapayaḥ& yati& yatayaḥ     \\
conjugation&ci& acīyethām& aś& āśyethām     \\
absolutive& dr & dīrtvā & aś & aśitvā      \\
infinitive&dr& daritum & aś & aśitum     \\
\hline
husband - wife&viṣṇuḥ& lakṣmī& rāmaḥ& sītā     \\
son - father& rāma& daśaratha& abhimanyu& arjuna\\
daughter - father&pārvatī& himavān& sītā& janaka\\
charioteer - warrior&dāruka& kṛṣṇa&  śalya& karṇa\\
defeated - victorious&kicaka& bhīma& rāvaṇa& rāma\\
son - mother&kṛṣṇaḥ&yaśodā&rāmaḥ&kauśalyā\\
\hline
\end{tabular}
    \caption{Examples of syntactic (top) and semantic (bottom) questions in our test set}
     \label{table:analogy_examples}

\end{table}
  
   To probe the information encoded in word embeddings, we build a comprehensive set of syntactic and semantic questions as illustrated in Table \ref{table:analogy_examples}. We use the Sanskrit Heritage Reader \cite{goyaldesign16,heut_13} for generating syntactic questions.  We consider the primary conjugation table for verbal root words and the declension table for nominal words for syntactic analogies. For conjugations, there are 10 families and popularly also called as \textit{ga\d{n}as}. Each \textit{ga\d{n}a} is further divided into \textit{parasmaipadī} and \textit{ātmanepadī}. Out of these 20 families, we choose 3 representative candidates to generate primary conjugations such that most of the inflectional variations are covered. Similarly,  nominal words are broadly divided into \textit{ajanta} and \textit{halanta}. Further, they are divided based on gender and the end of root words (for example, \textit{akārānta, ukārānta, nakārānta, jakāranta etc}). In this way, we consider 25 fine-grained families for nominals.  From each family, we select 3 representative words. Words occurring in same family display similar inflections variations. Therefore, to make questions, we strictly choose word pairs from the same family where the inflectional component belongs to the same position in the conjugation/declension table.
  For semantic questions,  we extract semantic relation from  \textit{Amarako\d{s}a} \cite{inproceedings_nair}, \textit{Rāmāyanam} and \textit{Mahābhāratam}. In semantic relationships, we consider six categories: husband - wife, son - father, daughter - father, charioteer - warrior, defeated - victorious, and son - mother. There are 6,415 semantic and 10,000 syntactic questions.  We only consider single token words in our test set.  Here, we use accuracy as a metric to measure the performance. We evaluate accuracy separately for semantic and syntactic questions.  

  \paragraph{Categorization:}
  As the name suggests, categorization is an evaluation task where the goal is to group nominal sets of concepts into natural categories \cite{baroni-etal-2014-dont}. For example, \textit{narmadā, godāvarī, kāverī} and \textit{gaṅgā} fall under the names of river category. This task is treated as an unsupervised clustering task where word representations are clustered into \textit{n} clusters (\textit{n} is the total number of gold standard categories). Here, performance is evaluated in terms of \textit{purity}. This metric represents the degree to which elements from the single gold standard category belong to the same cluster. If unsupervised clustering exactly replicates the gold standard clusters, then the purity score reaches $100\%$. Otherwise, it decreases based on the inability to replicate gold standard partitions. 
  
  Following \newcite{baroni-lenci-2010-distributional}, we construct a test set of 15 common categories picked from \textit{Amarako\d{s}a}. We exploit the ontological classification of \textit{Amarako\d{s}a} based on \textit{Vaiśeṣika}  ontology. 
 \textit{Amarako\d{s}a} is also marked with semantic relations  between words. These relations are mainly classified into hierarchical and associative. We use hierarchical relations such as hypernymy - hyponymy and holonymy - meronymy for this task.
  In each category, we select up to 10 concepts. Our categorization data contains the following categories: \textit{vādyopakaraṇam, vṛttiḥ, grahaḥ, devatā, paśuḥ, pakṣī, nadī, vṛkṣaḥ, upakaraṇam, ābharaṇam, alaukikasthānam, vāhanam, parvataḥ, vṛkṣaphalam} and \textit{puṣpam}. We also build a separate test set consisting of 45 syntactic categories where each category is made up of 25 concepts. Each category represents a particular morphological class. 

\section{Experiments}
\label{experiments}
\paragraph{Corpus:}
Our corpus contains texts from the Digital Corpus of Sanskrit (DCS), scraped data from Wikipedia and Vedabase corpus.\footnote{We use SLP1 transliteration scheme for training all the models.} The number of words in each section is 3.8 M, 1.7 M, and 0.2 M, respectively. DCS and Vedabase are segmented, but the Wikipedia data is unsegmented. We use this data for training all the embedding models. Most of the data in our corpus is in the form of poetry.

\paragraph{Hyper-parameters:} For tackling OOV problem faced by token-based embedding models such as \texttt{word2vec} and \texttt{GloVe}, we leverage \texttt{sentencepiece} tokenizer with 32,000 vocabulary size. We train \texttt{word2vec} and \texttt{GloVe} on \texttt{sentencepiece} segmented corpus. We employ following hyper-parameter settings for \texttt{word2vec} and \texttt{GloVe} : the embedding dimension size of 300, window size as 11, minimum word count for vocabulary as 1 and the number of epochs equal to 80. We keep all the remaining parameters same as used by original authors of the respective works. Next, we use raw sentences (instead of using  \texttt{sentencepiece} tokenized data) for training \texttt{FastText} and \texttt{CharLM}. We use following hyper-parameter setting for \texttt{FastText} : the embedding dimension of 300, window size of 11, the number of epochs 80, the minimum count for deciding vocabulary equals to 6, negative sampling loss with 5 negative samples, the learning rate as 0.025, minimum length of n-grams as 3 and maximum length of n-grams as 11. For \texttt{CharLM}, we use default parameters of small model architecture as \newcite{kim_charLM}.\footnote{\url{https://s3.amazonaws.com/models.huggingface.co/bert/albert-base-v2-config.json}} For \texttt{ELMo}, we use hyper-parameter settings of small model architecture.\footnote{\url{https://s3-us-west-2.amazonaws.com/allennlp/models/elmo/2x1024_128_2048cnn_1xhighway/elmo_2x1024_128_2048cnn_1xhighway_options.json}} For \texttt{ALBERT}, we use the same hyper-parameters as \texttt{albert-base-V2} model architecture except vocabulary size of 32,000 and the number of hidden layers equal to 6.

\paragraph{Results:} Table \ref{table:intrinsic_results} reports results on all the intrinsic tasks. We evaluate contextual models without context to assess how effective they are when compared with static models. Applying \textit{contextualized} models without context is similar to just using their underlying static word representation. \texttt{ELMo} outperforms all the models in all the intrinsic tasks with a large margin except the semantic categorization task. Surprisingly, with limited training data, the contextualized \texttt{ELMo} model achieved impressive overall performance compared to static models across all intrinsic tasks. The subword level segmentation with the \texttt{sentencepiece} model empowered the \texttt{word2vec} and \texttt{GloVe} to such an extent that they outperform \texttt{FastText} in analogy prediction tasks and on par in the remaining tasks. For resource-rich languages, \texttt{ALBERT} have shown the state of the art performance on many evaluation tasks. However, despite a thorough hyper-parameter search, the performance of \texttt{ALBERT} could not compete with static models. This may be due to a lack of massive training data. Similarly, except relatedness and syntactic categorization, \texttt{CharLM} model shows low performance with a large margin. 
\begin{table}[bht]
\begin{small}
    \centering
\resizebox{0.8\textwidth}{!}{%
 \begin{tabular}{ccccccccccc|ccc}
\toprule
 &\multicolumn{1}{c}{Relatedness} &\multicolumn{2}{c}{Categorization} &\multicolumn{1}{c}{Similarity} &\multicolumn{2}{c}{Analogy}\\
 \cmidrule(r){2-2}\cmidrule(l){3-4}\cmidrule(l){5-5}\cmidrule(l){6-7}
Model &f-score&syn    & sem &acc   & sym &sem  \\
 \cmidrule(r){2-2}\cmidrule(l){3-4}\cmidrule(l){5-5}\cmidrule(l){6-7}
word2vec & 30.90  & 0.22 & 0.41 & 37.34 & 30.65 & 11.01 \\
GloVe    & 31.50  & 0.25 & 0.39 & 37.41 & 30.04 & 14.01 \\
FastText & 31.20  & 0.14 & \textbf{0.47} & 37.87 & 16.65 & 6.01  \\
CharLM   & 35.21 & 0.33 & 0.28 & 35.94 & 2.11  & 0.03  \\
ELMo     & \textbf{37.17} & \textbf{0.86} & 0.41 & \textbf{42.15} & \textbf{56.80}  & \textbf{32.70}  \\
ALBERT   & 33.00    & 0.27 & 0.34 & 32.70  & 25.86 & 0.10 \\
\toprule
\end{tabular}}
    \caption{Results on intrinsic evaluation tasks. For similarity and analogy prediction tasks, we use accuracy (acc) in terms of percentage as the evaluation metric, purity for categorization and f-score for the relatedness task. The syntactic and semantic tasks are denoted by \textit{syn} and \textit{sem} keywords, respectively.}
    \label{table:intrinsic_results}
    \end{small}
\end{table}
\section{Analysis}
\subsection{Error Analysis}
We conduct a detailed analysis of prominent mistakes done by the word embedding models. For the analogy prediction task, we find that all the models fail to correctly predict analogy questions where the root word gets modified while adding the affix to the root word. For example, in this analogy question (\textit{lip : lepsyāmahe :: gur : \underline{goriṣyāmahe}}) while adding \textit{syāmahe} the root gets modified. On the other hand, all the models perform surprisingly well for analogy questions where affix is added without modifying root word (\textit{śak : śaknutam :: dagh : \underline{daghnutam}}). In the incorrect answers of the subword enriched \texttt{FastText} model, at least the root word is predicted correctly. \texttt{CharLM} performs the worst among all models and the predictions made by the model seem almost random with no relation to the root word or the suffix. \texttt{CharLM} learns a character level composition function to generate word-level representation. Therefore, it often predicts the correct root with an incorrect suffix or no suffix. While in the case of syntactic analogy prediction task, the \texttt{GloVe} and \texttt{word2vec} perform comparably. However, the improved performance of \texttt{GloVe} at semantic analogy task can be attributed to consideration of the global statistics of word occurrences. We observe that all the embedding models majorly get those test cases correct for similarity and relatedness tasks, which have words already present in the model vocabulary. This indicates that these models are only capable of generating representation for OOV words based on orthographic similarity and fail to capture semantic meanings. Specifically, around 55\% words present in the relatedness test set are OOV words.  

\subsection{Qualitative inspection for rare word}
\label{nearest_neighbour_analysis}
 \begin{table}[!h]
  \begin{small}
\centering
\begin{tabular}{ccccccc}
\cmidrule(r){1-7}
 &\multicolumn{3}{c}{\textbf{Vocabulary words}} & \multicolumn{3}{c}{\textbf{OOV words}}\\\cmidrule(r){2-4}\cmidrule(l){5-7}
         & \textit{nityam}      & \textit{sukham}        & \textit{tābhyām}       & \textit{ahham}      & \textit{bahūhuni}      & \textit{sahasra-cakṣo}     \\
         \cmidrule(r){2-4}\cmidrule(l){5-7} 
         
         & nityam-eva  & sukhamayaṃ    & āvābhyām      & allāh      & mānini        & pracakṣva         \\
word2vec  & nityamiti   & duḥkham       & dvābhyām      & ahne       & sani          & suparṇāḥ-ca       \\
         & sadā        & sukha-duḥkham & tayā          & ahrasat    & cūrṇitāni     & pracakṣmahe       \\
          \cmidrule(r){2-4}\cmidrule(l){5-7} 
         & nityamiti   & sukhamayaṃ    & adhikāribhyām & allāh      & bahū          & tat-ca-eva        \\
GloVe    & nityam-eva  & duḥkham       & pārṣṇibhyām   & ahne       & behulā        & tathā-ca          \\
         & sadā        & duḥkhameva    & jaṅghābhyām   & ahnuvi     & behulāyāḥ     & yat-ca-eva        \\
          \cmidrule(r){2-4}\cmidrule(l){5-7} 
         & sadā        & su-sukham     & tayoḥ         & simham     & bahuni        & sahasraḷaṃṇtra    \\
FastText & satatam     & duḥkham       & tau           & tam        & mandmatyoḥ    & sahasra-kṛtvas    \\
         & nitya-snāyī & sukha-duḥkham & cakṣuḍbhyām   & yathā-aham & bahavaḥ-tatra & sahasra-vedhī     \\
          \cmidrule(r){2-4}\cmidrule(l){5-7} 
         & sṭeḍiyam    & somam         & sabhāyāḥ      & īḍyam      & vatsyati      & paryaṭanasthaleṣu \\
CharLM   & samśayam    & saudham       & nyāyasabhāyāḥ & bṛhantam   & yotsyati      & naṣṭa-saṃjñaḥ     \\
         & yāpayan     & homam         & hastābhyām    & gayam      & setsyati      & cākṣuṣasya-antare \\
          \cmidrule(r){2-4}\cmidrule(l){5-7} 
         & martyam     & sakham        & nābhyām       & arham      & vilāsini      & ajāta-śatro       \\
ELMO     & niṣṭham     & kulam         & yābhyām       & aṅkam      & mattakāśini   & sahasra-śīrṣā     \\
         & yugyam      & sukhām        & bhūbhyām      & andham     & jānuni        & ahu-rūpaḥ         \\
          \cmidrule(r){2-4}\cmidrule(l){5-7} 
         & dānam       & satām         & ardham        & ahrasam    & ityadīni      & ūrdhva-ge         \\
ALBERT   & loke        & padam         & mandam        & allāh      & hrasāni       & sahasra-śīrṣā     \\
         & bhavitā     & kṣayam        & dhairyam      & ahnuthāḥ   & maṃdiram      & sarva-mahīkṣitām  \\
\hline
\end{tabular}
    \caption{Nearest neighbours (based on cosine similarity) for in vocabulary and  OOV words for the models trained on Sanskrit corpus. Last three words are OOV words. We use contextual models without a context. Applying \textit{contextualized} models without context is very similar to just using their underlying static word representation.}
     \label{table:in_vs_out_vocab}
     \end{small}
\end{table}
We analyze the word representations captured by different models. We consider \textit{contextualized} models also in this experiment. We use contextual models without a context to assess how effective are they when compared with static models. Applying \textit{contextualized} models without context is very similar to just using their underlying static word representation. For systematic investigation, we choose three representative words from following categories: (1) In Vocabulary words (2) Out of Vocabulary words. Table \ref{table:in_vs_out_vocab} illustrates the top three nearest neighbors based on cosine similarity between embeddings learned by different models. For frequently occurring words present in vocabulary all the embeddings except \texttt{CharLM} and \texttt{ALBERT} are able to capture lexical meaning. The quality of nearest neighbour candidates of \texttt{CharLM} and \texttt{ALBERT} is relatively poor compared to other models. We suspect that the possible reason for this behaviour may be due to less amount of training data.  The subword level modeling in \texttt{word2vec}, \texttt{GloVe} and \texttt{FastText} makes these models more inclined to find words which are orthographically similar. For these models, the representations seems to depend on surface string. For example, the nearest neighbours of \textit{nityam} and \textit{sukham} are very close in terms of edit distance. On the other hand, \texttt{ELMo} not only captures semantic level features but also orthographic features. This is very well evident from the following examples. The nearest neighbours produced by \texttt{ELMo} for \textit{nityam} and \textit{sukham} are lexically similar but orthographically different in terms of edit distance. However, the nearest neighbours of \textit{tābhyām} exhibits suffix similarity. The last three words are OOV where \textit{ahham}, \textit{bahūhuni} are words with incorrect spelling and \textit{sahasra-cakṣo} is the compound word. For OOV words also  similar trend holds. We generate representation for OOV for \texttt{word2vec} and \texttt{GloVe} using \texttt{sentencepiece} tokenizer. This tokenizer breaks OOV into subwords available in \texttt{sentencepiece} model vocabulary and representation of OOV word is sum of representations of these segmented subwords.

\subsection{Qualitative inspection of contextualized embeddings}
\label{context_cluster}
\begin{figure}[!h]
\centering
\subfloat[]{\includegraphics[width=3in]{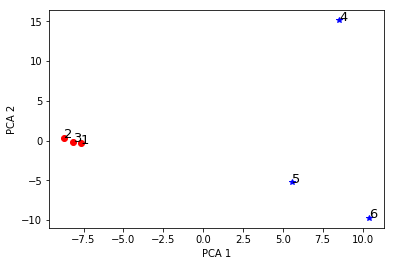}} 
\subfloat[]{\includegraphics[width=3in]{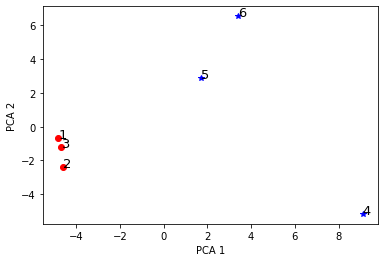}}\\
\caption{Projection of contextualized representation (the left side : \texttt{ELMo} and the right side \texttt{ALBERT}) for different sense of polysemous word \textit{cakrī} denoted by following sentences: \textit{(1) cakrī mṛdā jalena ca kumbhanirmāṇe viśāradaḥ
(2) cakrī kevala jala mṛd cakra upajīvakaḥ
(3) cakrī cakrasya upari mṛdaḥ kumbhaṃ nirmāti
(4) svahaste bibharti sudarśanam sa cakrī
(5) sudarśanena cakrī śiśupālam jaghāna
(6) rakṣobhyaḥ svabhaktān trātum cakrī sudarśanam bibharti.}} 
\label{fig:chakri} 
\end{figure}
\begin{figure}[!h]
\centering
\subfloat[]{\includegraphics[width=3in]{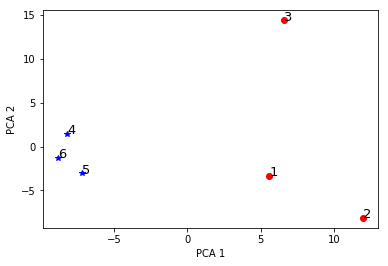}} 
\subfloat[]{\includegraphics[width=3in]{images/chakri_ALBERT.png}}\\
\caption{Projection of contextualized representation (the left side : \texttt{ELMo} and the right side \texttt{ALBERT}) for different sense of polysemous word \textit{hariḥ} denoted by following sentences: \textit{(1) vāyusutaḥ hariḥ vānarāṇām toṣam vavardha
(2) sugrīvaḥ nāma hariḥ vānarāṇām rājā āsīt
(3) vānarāṇām vṛddhaḥ hariḥ jāmbavān
(4) hariḥ baleḥ trailokyasampadam chalena jahāra
(5) hariḥ hiraṇyākṣam nihatya devatānām khedam jahāra
(6) hariḥ rukmiṇīm jahāra paśyatām}} 
\label{fig:harih} 
\end{figure}
To investigate the contextualized nature of \texttt{ELMo} and \texttt{ALBERT}, we analyze the vector representation (using PCA to project the vectors into 2D space) of a two polysemous words in different context. In this experiment, each test case consists of 6 sentences in Sanskrit such that all sentences have one common polysemous word. Out of 6 sentences, 3 sentences represent one sense of meaning and the remaining represent another sense of meaning for that common polysemous word. Ideally, contextualized word vectors representing the same sense of meaning should form a cluster. For the polysemous word \textit{cakrī}, we consider two senses of meaning, namely, potter and the Lord Viṣṇu. This test set consists of 6 sentences as follows: \textit{(1) cakrī mṛdā jalena ca kumbhanirmāṇe viśāradaḥ
(2) cakrī kevala jala mṛd cakra upajīvakaḥ
(3) cakrī cakrasya upari mṛdaḥ kumbhaṃ nirmāti
(4) svahaste bibharti sudarśanam sa cakrī
(5) sudarśanena cakrī śiśupālam jaghāna
(6) rakṣobhyaḥ svabhaktān trātum cakrī sudarśanam bibharti.} For the first three sentences, the sense of the meaning of the word \textit{cakrī} is a potter and for the remaining sentences, the sense of meaning is the Lord Viṣṇu.  Figure \ref{fig:chakri} illustrates visualization of polysemous word \textit{cakrī}. Here, we observe that the contextualized representation indicating the same sense of meaning are clustered together. Similarly, we consider another polysemous word \textit{hariḥ} consists of two senses of meaning, namely, monkey and the Lord Viṣṇu. \textit{(1) vāyusutaḥ hariḥ vānarāṇām toṣam vavardha
(2) sugrīvaḥ nāma hariḥ vānarāṇām rājā āsīt
(3) vānarāṇām vṛddhaḥ hariḥ jāmbavān
(4) hariḥ baleḥ trailokyasampadam chalena jahāra
(5) hariḥ hiraṇyākṣam nihatya devatānām khedam jahāra
(6) hariḥ rukmiṇīm jahāra paśyatām}. In Figure \ref{fig:harih}, we find the similar trend for this test case. The qualitative analysis suggests that the \textit{contextualized} representations generated by \newcite{elmo} and \newcite{ALBERT:} for polysemous words indicating the same sense of meaning forms a cluster for each sense of meaning (Section \ref{context_cluster}). This is indeed a serendipitous outcome for such data-hungry models when trained on negligibly small amount of corpus available for Sanskrit.


\section{Conclusion and future work}
In this work, we focused on evaluating word embedding approaches (initially proposed for resource-rich language English) for the Sanskrit language. 
Word embedding helps to transfer knowledge learned from readily available unlabelled data for improving the task-specific performance of data-driven approaches.
We investigat the effectiveness of word embedding approaches for Sanskrit. To facilitate systematic experimentation, we classify word embeddings in the broad categories and evaluated on 4 intrinsic tasks, namely, relatedness, categorization, similarity identification and analogy prediction tasks. This work can be considered as the fertile soil for investigating the following questions: (1) Which linguistic phenomenons are captured by word embeddings? (2) Out of existing word embedding approaches, which embeddings are more suitable for Sanskrit? (3) What are the shortcomings of existing approaches specific to Sanskrit? Surprisingly, with limited training data, the \newcite{elmo} achieved impressive overall performance compared to static models across all intrinsic tasks.  The qualitative analysis suggests that the \textit{contextualized} representations generated by \newcite{elmo} and \newcite{ALBERT:} for polysemous words indicating the same sense of meaning forms a cluster for each sense of meaning (Section \ref{context_cluster}). This is indeed a serendipitous outcome for such data-hungry models when trained on negligibly small amount of corpus available for Sanskrit.

There are many limitations to this study. While in this work, we restrict our study to intrinsic evaluation tasks for the standard word embedding approaches with their default setting. We plan to extend this work on extrinsic evaluation tasks as well. Moreover, to get substantial empirical evidence on linguistics phenomenons captured by these models, we plan to evaluate them for context-sensitive tasks similar to  \newcite{pilehvar-camacho-collados-2019-wic}. We find that all the models perform poorly on relatedness and similarity tasks since most of the words present in these evaluation test data are OOV words. However, these models perform relatively well for the words present in model vocabulary. This observation suggests investigating: What is the effect of the corpus size? We currently train our models with around 5 million tokens, which is negligible compared to the 840 billion tokens used by the resource-rich language models. 




\section*{Acknowledgements}
We are grateful to Amba Kulkarni for the discussion related to evaluation task decisions. We thank Mahesh Akavarapu for helping us with the data preparation step.
The TCS Fellowship supports the work of the first author under the Project Number  TCS/EE/2011191P.

\bibliographystyle{acl}
\bibliography{scl}
\end{document}